# ROBUST CLASSIFICATION WITH CONTEXT-SENSITIVE FEATURES

Peter D. Turney

Knowledge Systems Laboratory
Institute for Information Technology
National Research Council Canada
Ottawa, Ontario, Canada, K1A 0R6

turney@ai.iit.nrc.ca

**ABSTRACT**

This paper addresses the problem of classifying observations when features are context-sensitive, especially when the testing set involves a context that is different from the training set. The paper begins with a precise definition of the problem, then general strategies are presented for enhancing the performance of classification algorithms on this type of problem. These strategies are tested on three domains. The first domain is the diagnosis of gas turbine engines. The problem is to diagnose a faulty engine in one context, such as warm weather, when the fault has previously been seen only in another context, such as cold weather. The second domain is speech recognition. The context is given by the identity of the speaker. The problem is to recognize words spoken by a new speaker, not represented in the training set. The third domain is medical prognosis. The problem is to predict whether a patient with hepatitis will live or die. The context is the age of the patient. For all three domains, exploiting context results in substantially more accurate classification.

**INTRODUCTION**

A large body of research in machine learning is concerned with algorithms for classifying observations, where the observations are described by vectors in a multi-dimensional space of features. It often happens that a feature is context-sensitive. For example, when diagnosing spinal diseases, the significance of a certain level of flexibility in the spine depends on the age of the patient. This paper addresses the classification of observations when the features are context-sensitive.

In empirical studies of classification algorithms, it is common to randomly divide a set of data into a testing set and a training set. In this paper, for two of the three domains, the testing set and the training set have been deliberately chosen so that the contextual features range over values in the training set that are different from the values in the testing set. This adds an extra level of difficulty to the classification problem.

The paper begins with a precise definition of context. General strategies for exploiting contextual information are then given. The strategies are tested on three domains. First, the paper shows how contextual information can improve the diagnosis of faults in an aircraft gas turbine engine. The classification algorithms used on the engine data were a form of instance-based learning (IBL) [1, 2, 3] and a form of multivariate linear regression (MLR) [4]. Both algorithms benefit from contextual information. Second, the paper shows how context can be used to improve speech recognition. The speech recognition data were classified using IBL and cascade-correlation (CC) [5]. Again, both algorithms benefit from exploiting context. Third, the paper shows how context can be used to improve the accuracy of medical prognosis. Hepatitis data were classified using IBL.

The presentation of the results in the three test domains is followed by a discussion of the interpretation of the results. The work presented here is then compared with related work by other researchers and future work is discussed. Finally, the conclusion is given. For the three domains (engine diagnosis, speech recognition, and medical prognosis) and three classification algorithms (IBL, MLR, and CC) studied here, exploiting contextual information results in a significant increase in accuracy of classification.

## DEFINITION OF CONTEXT

This section presents a precise definition of context. Let $C$ be a finite set of classes. Let $F$ be an $n$-dimensional feature space. Let $\grave{x} = (x_0, x_1, ..., x_n)$ be a member of $C \times F$; that is, $(x_1, ..., x_n) \in F$ and $x_0 \in C$. We will use $\grave{x}$ to represent a variable and $\grave{a} = (a_0, a_1, ..., a_n)$ to represent a constant in $C \times F$. Let $p$ be a probability distribution defined on $C \times F$. In the definitions that follow, we will assume that $p$ is a discrete distribution. It is easy to extend these definitions for the continuous case.

**Primary Feature:** Feature $x_i$ (where $1 \leq i \leq n$) is a *primary feature* for predicting the class $x_0$ when there is a value $a_i$ of $x_i$ and there is a value $a_0$ of $x_0$ such that:

$$p(x_0 = a_0 | x_i = a_i) \neq p(x_0 = a_0) \quad (1)$$

In other words, the probability that $x_0 = a_0$, given $x_i = a_i$, is different from the probability that $x_0 = a_0$.

**Contextual Feature:** Feature $x_i$ (where $1 \leq i \leq n$) is a *contextual feature* for predicting the class $x_0$ when $x_i$ is *not* a primary feature for predicting the class $x_0$ and there is a value $\grave{a}$ of $\grave{x}$ such that:

$$p(x_0 = a_0 | x_1 = a_1, ..., x_n = a_n) \neq$$
$$p(x_0 = a_0 | x_1 = a_1, ..., x_{i-1} = a_{i-1}, \quad (2)$$
$$x_{i+1} = a_{i+1}, ..., x_n = a_n)$$

In other words, if $x_i$ is a contextual feature, then we can make a better prediction when we know the value $a_i$ of $x_i$ than we can make when the value is unknown, assuming that we know the values of the other features, $x_1, ..., x_{i-1}, x_{i+1}, ..., x_n$.

The definitions above refer to the class $x_0$. In the following, we will assume that the class is fixed, so that we do not need to explicitly mention the class.

**Irrelevant Feature:** Feature $x_i$ (where $1 \leq i \leq n$) is an *irrelevant feature* when $x_i$ is *neither* a primary feature *nor* a contextual feature.

**Context-Sensitive Feature:** A primary feature $x_i$ is *context-sensitive* to a contextual feature $x_j$ when there are values $a_0$, $a_i$, and $a_j$, such that:

$$p(x_0 = a_0 | x_i = a_i, x_j = a_j) \neq p(x_0 = a_0 | x_i = a_i) \quad (3)$$

The primary concern here is strategies for handling context-sensitive features.

Table 1 illustrates the above definitions. Since $p(x_0 = 1) = 0.5$ and $p(x_0 = 1 | x_1 = 1) = 0.44$, it follows that $x_1$ is a primary feature:

$$p(x_0 = 1) \neq p(x_0 = 1 | x_1 = 1) \quad (4)$$

Since $p(x_0 = a_0 | x_2 = a_2)$ equals $p(x_0 = a_0)$ for all values $a_0$ and $a_2$, it follows that $x_2$ is not a primary feature. However, $x_2$ is not an irrelevant feature, since:

$$p(x_0 = 1 | x_1 = 1, x_2 = 1, x_3 = 1)$$
$$\neq p(x_0 = 1 | x_1 = 1, x_3 = 1) \quad (5)$$

Therefore $x_2$ is a contextual feature. Furthermore, primary feature $x_1$ is context-sensitive to the contextual feature $x_2$, since:

$$p(x_0 = 1 | x_1 = 1, x_2 = 1) = 0.53 \quad (6)$$

and

$$p(x_0 = 1 | x_1 = 1) = 0.44 \quad (7)$$

Finally, $x_3$ is an irrelevant feature, since, for all values $a_0$, $a_1$, $a_2$, and $a_3$:

$$p(x_0 = a_0 | x_1 = a_1, x_2 = a_2, x_3 = a_3)$$
$$= p(x_0 = a_0 | x_1 = a_1, x_2 = a_2) \quad (8)$$

When $p$ is unknown, it is often possible to use background knowledge to distinguish primary, contextual, and irrelevant features. Examples of this use of background knowledge will be presented later in the paper.

Table 1: Examples of the different types of features.

| class | primary | contextual | irrelevant | probability |
|---|---|---|---|---|
| $x_0$ | $x_1$ | $x_2$ | $x_3$ | $p$ |
| 0 | 0 | 0 | 0 | 0.03 |
| 0 | 0 | 0 | 1 | 0.03 |
| 0 | 0 | 1 | 0 | 0.08 |
| 0 | 0 | 1 | 1 | 0.08 |
| 0 | 1 | 0 | 0 | 0.07 |
| 0 | 1 | 0 | 1 | 0.07 |
| 0 | 1 | 1 | 0 | 0.07 |
| 0 | 1 | 1 | 1 | 0.07 |
| 1 | 0 | 0 | 0 | 0.07 |
| 1 | 0 | 0 | 1 | 0.07 |
| 1 | 0 | 1 | 0 | 0.07 |
| 1 | 0 | 1 | 1 | 0.07 |
| 1 | 1 | 0 | 0 | 0.03 |
| 1 | 1 | 0 | 1 | 0.03 |
| 1 | 1 | 1 | 0 | 0.08 |
| 1 | 1 | 1 | 1 | 0.08 |

**STRATEGIES FOR EXPLOITING CONTEXT**

Katz *et al.* [6] list four strategies for using contextual information when classifying:

1. **Contextual normalization:** The contextual features can be used to normalize the context-sensitive primary features, prior to classification. The intent is to process context-sensitive features in a way that reduces their sensitivity to the context.

2. **Contextual expansion:** A feature space composed of primary features can be expanded with contextual features. The contextual features can be treated by the classifier in the same manner as the primary features.

3. **Contextual classifier selection:** Classification can proceed in two steps: First select a specialized classifier from a set of classifiers, based on the contextual features. Then apply the specialized classifier to the primary features.

4. **Contextual classification adjustment:** The two steps in strategy 3 can be reversed: First classify, using only the primary features. Then make an adjustment to the classification, based on the contextual features.

This paper examines strategies 1 and 2. A fifth strategy is also investigated:

5. **Contextual weighting:** The contextual features can be used to weight the primary features, prior to classification. The intent of weighting is to assign more importance to features that, in a given context, are more useful for classification.

The purpose of contextual normalization is to treat all features equally, by removing the affects of context and measurement scale. Contextual weighting has a different purpose: to prefer some features over other features, if they may improve accuracy.

**THE CLASSIFICATION ALGORITHMS**

To demonstrate the generality of the above strategies, three different classification algorithms were used, a form of instance-based learning (IBL) [1, 2, 3], multivariate linear regression (MLR) [4], and cascade-correlation (CC) [5].

Instance-based learning [1, 2] is closely related to the nearest neighbor pattern recognition paradigm [3]. Predictions are made by matching new data to stored data, using a measure of similarity to find the best matches [3]. The algorithm used here is a simple form of IBL, known as single-nearest neighbor pattern recognition. The algorithm is given, as input, an observation (a feature vector) in the testing set. To classify this observation, the algorithm simply looks for the most similar observation in the training set (the single nearest neighbor). The output of the algorithm is the class of the nearest neighbor in the training set. The measure of similarity used here is based on the sum of the absolute values of the differences between the elements of the vectors. If $\vec{x}$ and $\vec{y}$ are two feature vectors, then the similarity between $\vec{x}$ and $\vec{y}$ is defined as:

$$\sum_i (1 - |x_i - y_i|) \qquad (9)$$

In this paper, IBL will be used to refer to this simple form of instance-based learning. IBL easily handles both symbolic [1] and real-valued [2] features and classes.

Multivariate linear regression [4] models data with a system of linear equations. Like IBL, MLR easily handles both symbolic and real-valued features and classes. The algorithm used here is a form of MLR that is suitable for symbolic classes, known as linear discriminant analysis. In this paper, MLR will be used to refer to this form of multivariate linear regression. Suppose that there are $n$ distinct classes. In the training phase, MLR generates $n$ linear equations, one for each of the $n$ classes. The general form of the $n$ linear equations is:

$$y = \sum_i a_i x_i \qquad (10)$$

For example, consider the linear equation for one of the $n$ classes, class $j$. In the training phase, for each observation in the training set, $y$ is set to the value 1 when the observation belongs to class $j$. Otherwise, $y$ is set to 0. The $x_i$ in the equation for class $j$ are selected from among the features available in the feature space. MLR uses the forward selection procedure to select the $x_i$ [4]. Standard linear regression techniques are used to find the best values for the constant coefficients $a_i$ in the linear equation [7]. In the testing phase, MLR is given the values of the $x_i$ variables for each observation in the testing set. To predict the class of an observation, MLR calculates the values of the $n$ linear equations. This yields $n$ values for $y$, one value for each of the $n$ classes. The predicted class of the observation is the class with the largest calculated $y$ value.

Cascade-correlation (CC) [5] is a form of neural network algorithm. Like IBL and MLR, CC easily handles both symbolic and real-valued features and classes. The CC algorithm is similar to feed-forward neural networks trained with back-propagation. An interesting characteristic of the CC algorithm is that the network architecture is not specified by the user; it is determined automatically, by the CC algorithm. The network begins the training phase with no hidden-layer nodes. Hidden-layer nodes are then added, one-by-one, until a given performance criterion is met.

## GAS TURBINE ENGINE DIAGNOSIS

This section compares contextual normalization (strategy 1) with other popular forms of normalization. Strategies 2 to 5 are not examined in this section. The application is fault diagnosis of an aircraft gas turbine engine. The feature space consists of about 100 continuous primary features (engine performance parameters, such as thrust, fuel flow, and temperature) and 5 continuous contextual features (ambient weather conditions, such as external air temperature, barometric pressure, and humidity). The observations fall in eight classes: seven classes of deliberately implanted faults and a healthy class [7].

The amount of thrust produced by an engine is a primary feature for diagnosing faults in the engine. The exterior air temperature is a contextual feature, since the engine's performance is sensitive to the exterior air temperature. Exterior air temperature is not a primary feature, since knowing the exterior air temperature, *by itself*, does not help us to make a diagnosis. The experimental design ensures this, since the faults were deliberately implanted. This background knowledge lets us distinguish primary and contextual features, without having to determine the probability distribution.

The data consist of 242 observations, divided into two sets of roughly the same size. One set of observations was collected during warmer weather and the second set was collected during cooler weather. One set was used as the training set and the other as the testing set, then the sets were swapped and the process was repeated. Thus the sample size for testing purposes is 242.

The data were analyzed using two classification algorithms, IBL and MLR. IBL and MLR were also used to preprocess the data by contextual normalization [7].

The following methods for normalization were experimentally evaluated:

1. no normalization (use raw feature data)
2. normalization without context, using
   a. normalization by minimum and maximum value in the training set (the minimum is normalized to 0 and the maximum is normalized to 1)
   b. normalization by average and standard deviation in the training set (subtract the average and divide by the standard deviation)
   c. normalization by percentile in the training set (if 10% of the values of a feature are below a certain level, then that level is normalized to 0.1)
   d. normalization by average and standard deviation in a set of healthy baseline observations (chosen to span a range of ambient conditions)
3. contextual normalization (strategy 1), using
   a. IBL (trained with healthy baseline observations)
   b. MLR (trained with healthy baseline observations)

Of the five strategies for exploiting context, discussed above, only one was applied to the gas turbine engine data:

**Contextual normalization:** Let $\grave{x}$ be a vector of primary features and let $\grave{c}$ be a vector of contextual features. Contextual normalization transforms $\grave{x}$ to a vector $\grave{v}$ of normalized features, using the context $\grave{c}$. The following formula was used for contextual normalization:

$$v_i = \frac{x_i - \mu_i(\grave{c})}{\sigma_i(\grave{c})} \quad (11)$$

In (11), $\mu_i(\grave{c})$ is the expected value of $x_i$ and $\sigma_i(\grave{c})$ is the expected variation of $x_i$, as a function of the context $\grave{c}$. The values of $\mu_i(\grave{c})$ and $\sigma_i(\grave{c})$ were estimated using IBL and MLR, trained with healthy observations (spanning a range of ambient conditions) [7].

Table 2 (derived from Table 5 in [7]) shows the results of this experiment. For IBL, the average score without contextual normalization is 42% and the average score with contextual normalization is 55%, an improvement of 13%. For MLR, the average score without contextual normalization is 39% and the average score with contextual normalization is 46%, an improvement of 7%. According to the Student *t*-test, contextual normalization is signifi-cantly better than all of the alternatives that were examined [7].

Table 2: A comparison of various methods of normalization.

| classifier | normalization | no. correct (of 242) | percent correct |
|---|---|---|---|
| IBL | none | 102 | 42 |
| IBL | min/max train | 101 | 42 |
| IBL | avg/dev train | 97 | 40 |
| IBL | percentile train | 92 | 38 |
| IBL | avg/dev baseline | 111 | 46 |
| IBL | IBL | 139 | 57 |
| IBL | MLR | 128 | 53 |
| MLR | none | 100 | 41 |
| MLR | min/max train | 100 | 41 |
| MLR | avg/dev train | 100 | 41 |
| MLR | percentile train | 74 | 31 |
| MLR | avg/dev baseline | 100 | 41 |
| MLR | IBL | 103 | 43 |
| MLR | MLR | 119 | 49 |

**SPEECH RECOGNITION**

This section examines strategies 1, 2, and 5: contextual normalization, contextual expansion, and contextual weighting. The problem is to recognize a vowel spoken by an arbitrary speaker. There are ten continuous primary features (derived from spectral data) and two discrete contextual features (the speaker's identity and sex). The observations fall in eleven classes (eleven different vowels) [8].

For speech recognition, spectral data is a primary feature for recognizing a vowel. The sex of the speaker is a contextual feature, since we can achieve better recognition by exploiting the fact that a man's voice tends to sound different from a woman's voice. Sex is not a primary feature, since knowing the speaker's sex, *by itself*, does not help us to recognize a vowel. The experimental design ensures this, since all speakers spoke the same set of vowels. This background knowledge lets us distinguish primary and contextual features, without having to determine the probability distribution.

The data were divided into a training set and a testing set. Each of the eleven vowels was spoken six times by each speaker. The training set is from four male and four female speakers ($11 \times 6 \times 8 = 528$ observations). The testing set is from four new male and three new female speakers ($11 \times 6 \times 7 = 462$ observations). Using a wide variety of neural network algorithms, Robinson [9] achieved accuracies ranging from 33% to 56% correct on the testing set. The mean score was 49%, with a standard deviation of 6%. Table 3 summarizes Robinson's results.

Three of the five strategies discussed above were applied to the data:

**Contextual normalization:** Each feature was normalized by equation (11), where the context vector $\grave{c}$ was simply the speaker's identity. The values of $\mu_i(\grave{c})$ and $\sigma_i(\grave{c})$ were estimated simply by taking the average and standard deviation of $x_i$ for the speaker $\grave{c}$. In a practical application, this will require storing speech samples from a new speaker in a buffer, until enough data are collected to calculate the average and standard deviation.

**Contextual expansion:** The sex of the speaker was treated as another feature. This strategy is not applicable to the speaker's identity, since the speakers in the testing set are distinct from the speakers in the training set.

**Contextual weighting:** Let $\grave{x}$ be a vector of primary features and let $\grave{c}$ be a vector of contextual features. As with contextual normalization, the context vector $\grave{c}$ is simply the speaker's identity. The features were multiplied by weights, where the weight $w_i$ for a feature $x_i$ was the ratio of inter-class deviation $\sigma_i^{\text{inter}}$ to intra-class deviation $\sigma_i^{\text{intra}}$:

$$w_i = \frac{\sigma_i^{\text{inter}}}{\sigma_i^{\text{intra}}} \quad (12)$$

Table 3: Robinson's (1989) results with the vowel data.

| classifier | no. of hidden units | no. correct (of 462) | percent correct |
|---|---|---|---|
| Single-layer perceptron | - | 154 | 33 |
| Multi layer perceptron | 88 | 234 | 51 |
| Multi-layer perceptron | 22 | 206 | 45 |
| Multi-layer perceptron | 11 | 203 | 44 |
| Modified Kanerva Model | 528 | 231 | 50 |
| Modified Kanerva Model | 88 | 197 | 43 |
| Radial Basis Function | 528 | 247 | 53 |
| Radial Basis Function | 88 | 220 | 48 |
| Gaussian node network | 528 | 252 | 55 |
| Gaussian node network | 88 | 247 | 53 |
| Gaussian node network | 22 | 250 | 54 |
| Gaussian node network | 11 | 211 | 47 |
| Square node network | 88 | 253 | 55 |
| Square node network | 22 | 236 | 51 |
| Square node network | 11 | 217 | 50 |
| Nearest neighbor | - | 260 | 56 |

The inter-class deviation of a feature indicates the variation in a feature's value, across class boundaries. It is the average, for all speakers $\grave{c}$ in the training set, of the standard deviation of the feature, across all classes (all vowels), for a given speaker. Let $\sigma_1, ..., \sigma_m$ be the standard deviations of $x_i$ for each of the $m$ speakers in the training set. The inter-class deviation of $x_i$ is:

$$\sigma_i^{\text{inter}} = \frac{1}{m} \sum_{j=1}^{m} \sigma_j \quad (13)$$

The intra-class deviation of a feature indicates the variation in a feature's value, within a class boundary. It is the average, for all speakers in the training set and all classes, of the standard deviation of the feature, for a given speaker and a given class. Let $\{\sigma_{j,k}\}$, where $1 \leq j \leq m$ and $1 \leq k \leq n$, be the standard deviations of $x_i$ for each of the $m$ speakers and $n$ classes in the training set. The intra-class deviation of $x_i$ is:

$$\sigma_i^{intra} = \frac{1}{m}\frac{1}{n}\sum_{j=1}^{m}\sum_{k=1}^{n}\sigma_{j,k} \qquad (14)$$

The ratio of inter-class deviation to intra-class deviation is high when a feature varies greatly across class boundaries, but varies little within a class. A high weight (a high ratio) suggests that the feature will be useful for classification. This is a form of contextual weighting, because the weight is calculated on the basis of the speaker's identity, which is a contextual feature.

Table 4 shows the results of using different combinations of these three strategies with IBL. These results show that there is a form of synergy here, since the sum of the improvements of each strategy used separately is less than the improvement of the three strategies used together ($(58-56)+(55-56)+(58-56)=3\%$ for the sum of the three strategies used separately versus $66-56=10\%$ for the three strategies used together).

The three strategies were also tested with cascade-correlation [5]. Because of the time required for training CC, results were gathered for only two cases: With no preprocessing, cascade-correlation correctly classified 216 observations (47%). With preprocessing by all three strategies, cascade-correlation correctly classified 236 observations (51%). This shows that contextual information can be of benefit for both neural networks and nearest neighbor pattern recognition.

**HEPATITIS PROGNOSIS**

Similar to the previous section, this section examines strategies 1, 2, and 5: contextual normalization, contextual expansion, and contextual weighting. The problem is to determine whether hepatitis patients will live or die from their disease. There are seventeen primary features, of which twelve are discrete (such as "patient is taking steroids", "patient reports fatigue") and five are continuous (such as "patient's bilirubin level"). There are two contextual features, of which one is discrete (patient's sex) and one is continuous (patient's age). The patient's sex was not used in the following experiments, since 90% of the patients were male. The observations fall in two classes (live or die) [10]. There are many missing values in the hepatitis data. These were filled in by using the single-nearest neighbor algorithm with the training data.

For hepatitis prognosis, bilirubin level is a primary feature for determining whether the patient will die from the disease. The age of the patient is a contextual feature, since we can achieve more accurate prognoses by using the patient's age. Age is not a primary feature, since knowing the patient's age, *by itself*, does not help us to make a prognosis. In support of this claim, compare rows one and three in Table 5. Adding age as a feature actually *reduces* accuracy. Background knowledge does not help us to determine whether age is primary or contextual, since it is plausible that the patient's age could be a primary factor in hepatitis prognosis. In this case, we must use the data to estimate the probability distribution. The data suggest that age is a contextual feature.

The data were divided into a training set and a testing set. Unlike the previous two experiments, there was no systematic distinction between the training and testing sets. The data consist of 155 observations, which were randomly split to make 10 pairs of training and testing sets. In each pair, there were 100 training observations and 55 testing observations. Thus the total number of observations for testing purposes was 550.

Three of the five strategies discussed above were applied to the data:

**Contextual normalization:** Each feature was normalized by equation (11), where the context vector $\vec{c}$ is simply the patient's age. Age was converted into a discrete feature by dividing age into five intervals, with an equal number of patients in each interval. The values of $\mu_i(\grave{c})$ and $\sigma_i(\grave{c})$ were estimated by taking the average and standard deviation of $x_i$ for each interval $\grave{c}$. This is different from the method used for contextual normalization with the continuous contextual features in gas turbine engine diagnosis [7]. Note that equation (11) does not require continuous features; it works well with the boolean features in the hepatitis data, when true and false are represented by one and zero.

Table 4: The three strategies applied to the vowel data.

| strategy 1: contextual normalization | strategy 2: contextual expansion | strategy 5: contextual weighting | no. correct (of 462) | percent correct |
|---|---|---|---|---|
| No | No | No | 258 | 56 |
| No | No | Yes | 269 | 58 |
| No | Yes | No | 253 | 55 |
| No | Yes | Yes | 272 | 59 |
| Yes | No | No | 267 | 58 |
| Yes | No | Yes | 295 | 64 |
| Yes | Yes | No | 273 | 59 |
| Yes | Yes | Yes | 305 | 66 |

Table 5: The three strategies applied to the hepatitis data.

| strategy 1: contextual normalization | strategy 2: contextual expansion | strategy 5: contextual weighting | no. correct (of 550) | percent correct |
|---|---|---|---|---|
| No | No | No | 393 | 71 |
| No | No | Yes | 393 | 71 |
| No | Yes | No | 390 | 71 |
| No | Yes | Yes | 391 | 71 |
| Yes | No | No | 454 | 83 |
| Yes | No | Yes | 460 | 84 |
| Yes | Yes | No | 457 | 83 |
| Yes | Yes | Yes | 464 | 84 |

**Contextual expansion:** The age of the patient was treated as another feature. This strategy is not useful for the patient's sex, since so few patients are female.

**Contextual weighting:** The features were multiplied by weights, where the weight for a feature was the ratio of inter-class deviation to intra-class deviation, as in equation (12). The inter-class deviation and the intra-class deviation were calculated using the five age intervals.

Table 5 shows the results of using different combinations of the three strategies (contextual normalization, contextual expansion, and contextual weighting) with IBL. As in the previous section, there is a form of synergy here, since the sum of the improvements of each strategy used separately is less than the improvement of the three strategies together ( $(71 - 71)$ + $(71 - 71)$ + $(83 - 71)$ = 12% for the sum of the three strategies versus $84 - 71 = 13\%$ for the three strategies used together). In this case, however, the synergy is not as marked as it is in the previous section. This may be due to the fact that there is no systematic difference between the training and testing sets in the hepatitis data, while the testing set for the vowel data uses different speakers from the training set.

For comparison, other researchers have reported accuracies of 80% [11] and 83% [12] on the hepatitis data. It is interesting that a single-nearest neighbor algorithm can match or surpass these results, when strategies are employed to use the contextual information contained in the data.

**DISCUSSION OF RESULTS**

The results reported above indicate that contextual normalization and contextual weighting can significantly improve the accuracy of classification. Contextual expansion is less effective than contextual normalization and contextual weighting, although it appears useful, when used in conjunction with the other techniques.

Equation (11) (a form of contextual normalization) has three characteristics:

1. The normalized features all have the same scale, so we can directly compare features that were originally measured with different scales.

2. Equation (11) tends to weight features according to their relevance for classification. Features that are far from average, in a given context, are normalized to values that are far from zero. That is, a surprising feature will get a high absolute value. A feature that is irrelevant will tend to have a high variation, so it will tend to be normalized to a value near zero. A feature that is near average will also be normalized to a value near zero. Note that this is true for boolean features, as well as continuous features.

3. Equation (11) compensates for variations in a feature that are due to variations in the context. Thus it reduces the impact of the context, allowing the classification system to generalize across different contexts more easily.

Equation (11) is only one possible form of contextual normalization. For example, another form of contextual normalization could use a context-sensitive estimate of the minimum and maximum values to normalize a feature.

Contextual weighting is a new technique for using contextual information. The idea of contextual weighting is to assign more weight to the features that seem more useful for classification, in a given context. Equation (12) is only one possible form of contextual weighting. For example, another form of contextual weighting might vary the weight as a function of the context. With equation (12), the weight is calculated using contextual information, but the weight does not change as a function of the context.

Note that equation (11) is a linear transformation of the data when the context $\grave{c}$ is constant, but it is a nonlinear transformation when the context is variable. Equation (12) is a linear transformation of the data, both when the context $\grave{c}$ is constant and when it is variable, since the weight $w_i$ is fixed; it does not vary with the context $\grave{c}$.

Of the three classification algorithms, IBL gained the most from contextual normalization and contextual weighting. The form of IBL that was used here (single-nearest neighbor with sum of absolute values as a distance measure) is particularly sensitive to the scales of the features. If one feature ranges from 0 to 100 and the remaining features range from 0 to 1, then the first feature will have much more influence on the distance measure than the remaining features. Therefore IBL can benefit significantly from contextual normalization, which attempts to equalize scales. MLR and CC are designed to be unaffected by linear transformations of the features. Therefore they do not favor features with larger ranges. However, this strength is also a weakness, because MLR and CC cannot benefit from preprocessing of the data that increases the scale of more significant variables. For example, contextual weighting (using equation (12)) has no effect on MLR and it has only minor effects on CC.

It seems natural that contextual normalization and contextual weighting combine synergistically. Raw data consist of features that have essentially random scales. The scale of a feature usually has no relation to the importance of the feature for classification. Contextual normalization adjusts the features so that their scales are more equal. It seems plausible that, in many cases, assigning equal scales to the features is better for classification than assigning random scales to the features. Contextual weighting emphasizes the features that are most relevant for classification. Again, it seems plausible that, in many cases, contextual weighting will work better when the features have first been adjusted, so that they have equal scales. Thus the synergy found in the experiments reported here is to be expected.

**RELATED WORK**

The work described here is most closely related to [6]. However, [6] did not give a precise definition of the distinction between contextual features (their terminology: parameters or global features) and primary features (their terminology: features). They examined only contextual classifier selection, using neural networks to classify images, with context such as lighting. They found that contextual classifier selection resulted in increased accuracy and efficiency. They did not address the difficulties that arise when the context in the testing set is different from the context in the training set.

This work is also related to work in speech recognition on speaker normalization [8]. However, the work on speaker normalization tends to be specific to speech recognition. Here, the concern is with general-purpose strategies for exploiting context.

**FUTURE WORK**

Future work will extend the list of strategies, the list of domains that have been examined, and the list of classification algorithms that have been tested. It may also be possible and interesting to develop a general theory of strategies for exploiting context.

Due to its simplicity, IBL can easily be enhanced with strategies for exploiting context. Other classification algorithms can also be enhanced, but it may require more effort. It should be possible to modify algorithms such as MLR and CC so that they can benefit from a form of contextual weighting. For example, instead of preprocessing the data by multiplying the features by weights, a classification algorithm can be designed to take the original data and the set of weights as two separate sets of inputs. The algorithm can then use the weights to adjust its internal processing of the original data. MLR could use the contextual weights to decide which features it should include in its linear equations.

Another possibility is to design classification algorithms that can automatically distinguish primary features from contextual features. The definitions given in equations (1) and (2) should allow automatic distinction.

**CONCLUSIONS**

The general problem examined here is to accurately classify observations that have context-sensitive features. Examples are: the diagnosis of spinal problems, given that spinal tests are sensitive to the age of the patient; the diagnosis of gas turbine engine faults, given that engine performance is sensitive to ambient weather conditions; the recognition of speech, given that different speakers have different voices; the prognosis of hepatitis, given the patient's age; the classification of images, given varying lighting conditions. There is clearly a need for general strategies for exploiting contextual information. The results presented here demonstrate that contextual information can be used to increase the accuracy of classifiers, particularly when the context in the testing set is different from the context in the training set.

**ACKNOWLEDGMENTS**

The gas turbine engine data and engine expertise were provided by the Engine Laboratory of the NRC, with funding from DND. The vowel data and the hepatitis data were obtained from the University of California data repository (ftp ics.uci.edu, directory /pub/machine-learning-databases) [10]. The cascade-correlation [5] software was obtained from Carnegie-Mellon University (ftp pt.cs.cmu.edu, directory /afs/cs/project/connect/code). The author wishes to thank Rob Wylie and Peter Clark of the NRC and two anonymous referees of IEA/AIE-93 for their helpful comments on this paper.

This paper is an expanded version of a paper that first appeared in the *Proceedings of the European Conference on Machine Learning, 1993*. The author wishes to thank the conference chairs of both IEA/AIE-93 and ECML-93 for permitting this paper to appear here.